\documentclass[fleqn,10pt]{olplainarticle}

\title{Climate Intervention Analysis using AI Model Guided by Statistical Physics Principles}

\author[1]{Soo Kyung Kim}
\author[1]{Kalai Ramea}
\author[3]{Salva Rühling Cachay}
\author[2]{Haruki Hirasawa}
\author[1]{Subhashis Hazarika}
\author[2]{Dipti Hingmire}
\author[4]{Peetak Mitra}
\author[5]{Philip J. Rasch}
\author[2]{Hansi A. Singh}
\affil[1]{Palo Alto Research Center}
\affil[2]{University of Victoria}
\affil[3]{University of California, San Diego}
\affil[4]{Excarta}
\affil[5]{University of Washington}

\keywords{Fluctuation-Dissipation Theorem, Physics guided ML, Climate Change, Climate Intervention modeling}

\begin{abstract}
The availability of training data remains a significant obstacle for the implementation of machine learning in scientific applications. In particular, estimating how a system might respond to external forcings or perturbations requires specialized labeled data or targeted simulations, which may be computationally intensive to generate at scale. In this study, we propose a novel solution to this challenge by utilizing a principle from statistical physics known as the Fluctuation-Dissipation Theorem (FDT) to discover knowledge using an AI model that can rapidly produce scenarios for different external forcings. By leveraging FDT, we are able to extract information encoded in a large dataset produced by Earth System Models, which includes 8250 years of internal climate fluctuations, to estimate the climate system's response to forcings. Our model, AiBEDO, is capable of capturing the complex, multi-timescale effects of radiation perturbations on global and regional surface climate, allowing for a substantial acceleration of the exploration of the impacts of spatially-heterogenous climate forcers. To demonstrate the utility of AiBEDO, we use the example of a climate intervention technique called Marine Cloud Brightening, with the ultimate goal of optimizing the spatial pattern of cloud brightening to achieve regional climate targets and prevent known climate tipping points. While we showcase the effectiveness of our approach in the context of climate science, it is generally applicable to other scientific disciplines that are limited by the extensive computational demands of domain simulation models. Source code of AiBEDO framework is made available at \url{https://github.com/kramea/kdd_aibedo}. A sample dataset is made available at \url{https://doi.org/10.5281/zenodo.7597027}. Additional data available upon request. 
\end{abstract}

\begin{document}

\flushbottom
\maketitle
\thispagestyle{empty}

\section{Introduction}

Machine learning has greatly benefited a variety of scientific disciplines through its ability to extract valuable insights from large datasets. Notable examples include the use of deep learning for numerical weather prediction~\cite{bochenek2022machine, ren2021deep, markovics2022comparison}, medical imaging diagnosis~\cite{shen2017deep, razzak2018deep, erickson2017machine, chan2020deep, litjens2017survey}, classification of astronomical images~\cite{carrasco2019deep, hansen1984climate}, and drug discovery with specific properties~\cite{chen2018rise, stephenson2019survey, dara2022machine, patel2020machine}. These applications involve direct application of machine learning algorithms on large datasets in order to achieve a specific goal. Additionally, we have seen a rise in the use of surrogate models, such as data-driven models trained on simulation data, which can instantaneously reproduce output for a given set of inputs, thereby significantly increasing process efficiency~\cite{barkanyi2021modelling}. Despite these achievements, these use cases still represent only a small subset of problems studied by scientists. 

A commonly researched topic in systems analysis, science, and engineering is how a system responds to external forcing. This can encompass a wide range of questions, such as how a bridge or aircraft responds to wind~\cite{matsumoto1995response, siringoringo2012observed}, how a certain material responds to electrical conductivity or magnetic forces~\cite{reyne1987survey,zukoski1993material}, or how a financial market responds to geopolitical instability~\cite{hoque2020global,sohag2022stock}. As these inquiries often fall outside of the training data distributions, they are typically addressed through targeted simulation scenario runs. However, the complexity of the domain simulation model can significantly impact the computational time required for a single simulation run, ranging from a few hours to several months, which can impede the ability to rapidly evaluate "what-if" scenarios and gain insights into the system's response to various external forcings.

This limitation is particularly acute in climate science due to the high computational cost of Earth System Models (ESM). ESMs are comprehensive simulation models that integrate the physics of Earth's myriad processes to improve our understanding of the interactions between the atmosphere, oceans, land surface, and ice. These models are used for studying future changes in the Earth's climate~\cite{flato2011earth, giorgi2018regional}. The related concept of \textit{climate intervention} refers to proposed methods aimed at reducing the effects of climate change by manipulating the Earth's environment~\cite{mah2020coping, lee2019systematic, valentini2006goal}. One key category of intervention strategies is Solar Radiation Management~\cite{robock2020benefits, nicholson2018solar}, which involves reflecting sunlight back into space to cool the Earth. There are two main approaches that have been proposed to increase the reflection of sunlight on planetary scales: (a) Stratospheric Aerosol Injection (SAI)~\cite{ pope2012stratospheric,hulme2012climate} and (b) Marine Cloud Brightening (MCB)~\cite{ stjern_response_2018,latham2012marine,latham2012weakening}. SAI involves the introduction of aerosol particles such as sulphates into the stratosphere, while MCB involves the spraying of sea salt aerosols into low clouds in the lower atmosphere. While SAI generally has similar climate impacts irrespective of injection site, the effects of MCB are regional and spatially heterogenous. Both techniques necessitate extensive research and an in-depth understanding of their impacts on the Earth System, such as identifying possible unintended consequences of interventions. Moreover, because MCB is applied regionally, an immense number of different intervention scenarios are possible, making it an ideal use case for a rapid screening tool to evaluate efficacy before running costly ESM simulations.

Due to the complexity of modern ESMs, a single ESM simulation can take weeks to months to obtain sufficient sample sizes for analysis. Thus, past evaluations of SAI and MCB interventions have only considered a small number of scenarios, which are often involve simplified interventions \cite{kravitz2015geoengineering}. This is particularly true of MCB, where only a handful of cloud perturbation scenarios have been considered \cite{rasch_geoengineering_2009,jones_climate_2009,stjern2018response}.  As a result, a complete set of ESM simulation runs for MCB that can be used to query a number of ``what-if" scenarios does not exist. Given the large number of possible combinations of spatial and temporal cloud perturbations, studying them using ESMs is likely a computationally intractable task. While machine learning has been used in a variety of climate modeling contexts, such as improving and accelerating ESMs through the use of data-driven and hybrid physics-informed surrogate models~\cite{reichstein2019deep, ardabili2020deep, zhu2022building, lu2019efficient, weber2020deep}, the problem of determining the outcome of an external forcing (i.e., the outcome in a system due to an external cloud perturbation in the case of MCB) remains a challenge as it is a complex problem with limited data sets to train on. This makes direct application of machine learning approaches difficult, and requires a creative solution.  

We present AiBEDO, an AI model, rooted in an innovative application of the Fluctuation-Dissipation Theorem (FDT). FDT is a fundamental principle in statistical physics that states that the forced response of a system mirrors its internal fluctuations~\cite{kubo_fdt_1966}. FDT has been successfully used in many fields--brownian motion and drag in moving objects~\cite{tsekov2010brownian,sharma2004direct}, protein folding in biological systems~\cite{hayashi2007violation, mo2006quantum, huang2008protein} and assessing climate sensitivity in ESMs~\cite{leith_climate_1975,gritsoun2002construction, cionni_fluctuation_2004, dymnikov2005current, gritsun2007climate}. Past implementations of FDT in the scientific machine learning community mainly involve utilizing it to ensure the consistency of the physical dynamics of a system. Recently, Lee et al. (2021)~\cite{lee2021machine} apply FDT in machine learning to ensure thermodynamic consistency while learning reversible and irreversible dynamics of physical systems. A related but distinct study illustrates the conservation of the stationary equilibrium of parameters of the ML model using FDT by connecting measurable quantities and hyperparameters in the stochastic gradient descent algorithm~\cite{yaida2018fluctuation}.

Our use of FDT in machine learning differs from the previous work and is grounded in the fundamental application of learning from internal variability, intrinsic chaotic fluctuations in a complex system, to estimate a forced response. Performing this operation in the traditional modeling sense requires large sample sizes and is limited by the use of linear response functions. Our work illustrates a novel pathway to improve this baseline through the help of AI models. To achieve this goal, we train the model on the climate response over a wide range of natural fluctuations in the cloud radiative forcing~\cite{ramanathan1989cloud}. The training data is derived from a large ensemble of ESM simulations, amounting to a total of over 100,000 model months. The model can then estimate responses to forced changes in radiation, obviating the need to run many simulation runs and saving millions of core hours of computing resources. We validate the model and verify that it can plausibly project climate responses by comparing it with targeted ESM runs with regional MCB-like forcing. Key contribution of our work are listed as follows:
\begin{itemize}

    \item We have showcased a \textbf{novel method for developing a scenario analysis tool using the principle of Fluctuation-Dissipation Theorem to develop an AI model trained on noise} to answer scientific queries requiring simulation runs with external forcings. 

    \item Our AI framework maps the relationship between cloud perturbations and the climate response, including large-scale circulation and regional climate patterns. To the best of our knowledge, \textbf{this is a pioneering application of AI for scenario analysis of global-scale marine cloud brightening climate interventions}. 

    \item We have extensively evaluated the performance of AiBEDO against targeted simulation runs of ESMs and show that our model can reproduce the results with \textbf{high fidelity, but three orders of magnitude faster than ESMs}.

    \item We have developed an interactive post-hoc analysis platform to \textbf{examine the results and facilitate rapid prototyping of MCB what-if scenarios} for downstream decision-making. 

\end{itemize}

The remainder of the paper is organized as follows. Section 2 introduces the problem statement and details of how FDT is applied in AiBEDO. In Section 3, we see the details of different components of the MCB climate intervention framework. In Section 4, we evaluate the performance of the data-driven emulation component of AiBEDO against ESM runs, and we demonstrate the utility of AiBEDO in an FDT-like approach to perform MCB intervention scenarios. Finally, in Section 5, we discuss our results, limitations, and future directions. 

\section{Problem Statement}

The goal is to create a framework that one can use to rapidly generate MCB intervention impacts for a given spatial and temporal extent of cloud perturbations (external forcing). The problem will be addressed in two phases. In the first phase, we develop an AI emulator to map relationships between a set of designated input and output variables. This will involve the creation of a series of mappings that are performed at different time-lagged intervals. In the second phase, we sum them using an FDT operator to obtain a time-integrated outcome, which estimates the regional impact of the external forcing. 

\begin{table*}[t]
\centering
\begin{tabular}{|l|l|l|}
    \hline
    \textbf{Variable} & \textbf{Description} & \textbf{Role in AiBEDO} \\
    \hline
    \texttt{cres} & Net TOA shortwave cloud radiative effect & input \\ \hline
    \texttt{crel} & Net TOA longwave cloud radiative effect & input \\ \hline
    \texttt{cresSurf} & Net Surface shortwave cloud radiative effect & input \\ \hline
    \texttt{crelSurf} & Net Surface longwave cloud radiative effect & input \\ \hline
    \texttt{netTOAcs} & Net TOA clear-sky radiative flux & input \\ \hline
    \texttt{netSurfcs} & Net surface clear-sky radiative flux plus all-sky surface heat flux & input \\ \hline
    \texttt{lsMask} & Land fraction & input \\ \hline\hline
    \texttt{ps} & Surface pressure & output \\ \hline
    \texttt{tas} & Surface air temperature & output \\ \hline
    \texttt{pr} & Precipitation & output \\ \hline
\end{tabular}
\caption{CESM2 LE variables used in AiBEDO. TOA - top of atmosphere. All radiative and heat fluxes are positive down}
\label{tab:variables}
\end{table*}

\subsection{Phase 1: AI Emulator of simultaneous and time-lagged mappings}

Let us denote an input field of cloud and clearsky radiation anomalies at time $t$ as $\vec{x}(t)$ and denote the corresponding output field of surface climate anomalies after a time delay of $\tau$ as $\vec{y}(t+\tau)$. Our task is then to develop a model $A_{\tau}$ to predict $y(t+\tau)$ from $\vec{x}(t)$, i.e, $\vec{y}(t + \tau)  = A_{\tau} (\vec{x}(t))$. Formally, the input tensor $\delta \vec{x}(t)  \in \mathbb{R}^{d\times c_{in}}$, and output tensor $\vec{y}(t+\tau) \in \mathbb{R}^{d\times c_{out}}$ consist of 1-D climate data of size $d$ with corresponding channels of size $c_{in}$ and $c_{out}$. Input channels, $c_{in}$, are composed of 7 climate variables empirically selected to achieve best performance of the model $A_\tau$, and output channels, $c_{out}$, are composed of 7 climate variables. The list of climate variables used for input and output is shown in Table~\ref{tab:variables}. 

\subsection{Phase 2: Time-integrated output using Fluctuation-Dissipation theorem }

One common method of applying FDT involves assuming the metrics of interest have Gaussian statistics and constructing an FDT operator $\mathbf{L}$ using the covariance matrix $\mathbf{C}(t)$ \cite{cionni_fluctuation_2004,liu_sensitivity_2018}. The climate mean response $\delta \left<y\right>$ to a constant forcing $\delta \vec{f}$ is then computed as
\begin{equation}
    \left<\delta y\right> = \mathbf{L}^{-1} \delta \vec{f} = \left[\int_0^\infty \mathbf{C}(\tau) \mathbf{C}(0)^{-1} d\tau \right] \delta \vec{f}
\end{equation}

Noting that FDT is limited to the linear component of the climate response, we use an AI model with the intention of capturing non-linear components of the response and loosening some of the conditions required by classical FDT (namely that the probability density function of the relevant climate statistics must be Gaussian or quasi-Gaussian \cite{cionni_fluctuation_2004,majda_high_2010}). Thus, replacing the linear FDT operator $\mathbf{L}$ with a set of AiBEDO operators, $A_{\tau}$, at a series of time-lags ($\tau$), we construct an estimate of the response as
\begin{equation}
    \left<{\delta \vec{y}(t)}\right> = \sum_{\tau=0}^{T_{max}} \frac{1}{N}\sum_{i=0}^N{\left(A_{\tau}(\vec{x_i} + \delta \vec{f}(t-\tau)) - A_{\tau}(\vec{x_i})\right)}
    \label{eq:fdt_aibedo}
\end{equation}
where $\vec{x_i}$ are randomly sampled internal fluctuations of the input variables, $N$ is the number of samples of internal fluctuations used, and $T_{max}$ is the upper lag limit set at the point when the response to a perturbation approximately converges to noise. For testing purposes we use the first 6 months and for the full lag integration we choose 48 months.

\section{MCB Climate Intervention using AiBEDO: Data and Methods}

\begin{figure}[H]
\begin{center}
\includegraphics[width=7.5cm]{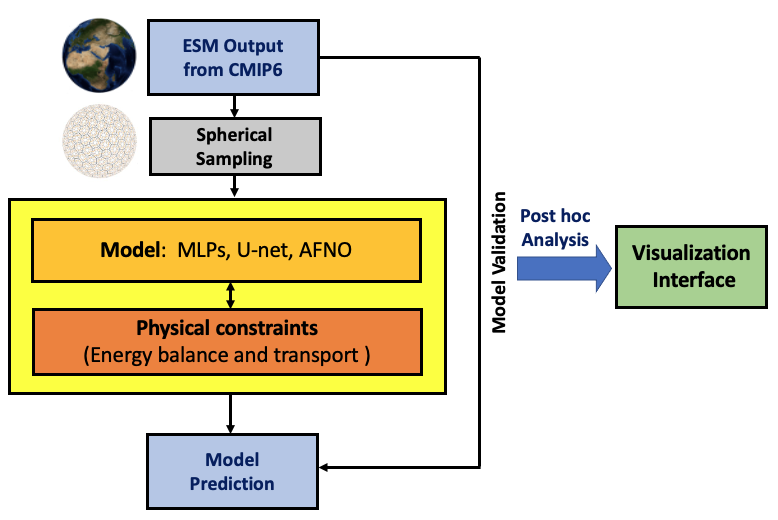}
\end{center}
\caption{Schematic view of MCB Climate Intervention framework}
\label{fig:aibedo}
\end{figure}

The MCB climate intervention framework consists of three main components: (1) datasets for training and verification, which includes a large ensemble of Earth System Model output data; (2) AiBEDO, which serves as the framework's central component is an AI model incorporating data-driven models; (3) an interactive visualization interface featuring dual functionality allowing users to execute rapid ``what-if'' scenarios and enabling modelers to inspect and provide explanations for the associated outcomes. A schematic of the AiBEDO framework is shown in Figure~\ref{fig:aibedo}.

\subsection{Data}
Our training data consists of a subset of the most recent sixth generation of Coupled Model Intercomparison (CMIP) ESM outputs. As FDT requires a large amount of climate noise (i.e. internal variability or chaotic fluctuations) for training, we use the Community Earth System Model 2 Large Ensemble (CESM2-LE) as a source for internal climate noise \cite[][; Table~\ref{tab:cesm2}]{rodgers_ubiquity_2021}. Specifically, we use the 50-member ensemble of historical simulations with smoothed biomass burning emission between 1997 and 2014 at nominal 1 degree spatial resolution and at a monthly temporal resolution. Each simulation the large ensemble is identical to one others except in their initial conditions, meaning they differ only in the chaotic fluctuations internal to the climate system. Thus, this provides one of the largest data sets of simulations from a single CMIP6-generation ESM from which we can obtain internal variability to train and test our model on, as it provides a total of nearly 100,000 months of data. Additionally, we use a set of novel CESM2 simulations which estimate the climate response to MCB-like perturbations (described in Appendix \ref{appendix:CESM2_MCB}. These are used to evaluate AiBEDO's ability to estimate the climate response to a forcing.

\subsubsection{Spherical Sampling}

\begin{figure}[H]
\begin{center}
\includegraphics[width=8.5cm]{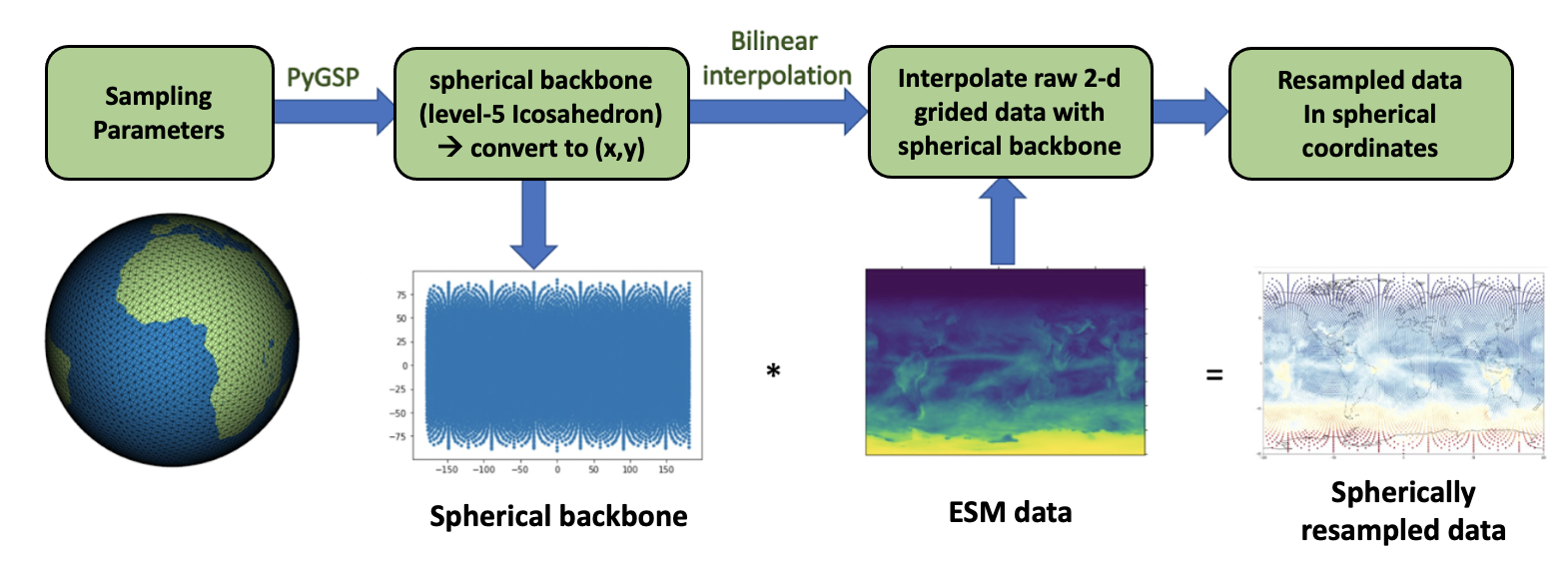}
\end{center}
\caption{Spherical sampling of ESM data following icosahedral grid.}
\label{fig:spherical_sampling}
\end{figure}

Earth System Model data is typically stored on 2D latitude-longitude gridded meshes, which have non-uniform area over the globe with smaller areas at the poles and larger areas at the equator, complicating their use in ML. Specifically, the rotational symmetry of the Earth is difficult to represent using two-dimensional meshes, resulting in inaccurate representations of significant climate patterns in ML models that assume a two-dimensional format of data, such as Convolutional Neural Networks (CNNs) or U-Nets. To address this limitation, we have adopted a geodesy-aware spherical sampling technique that converts the 2D rectangular grid to a 1D spherical icosahedral mesh, following the strategy suggested in the recent work~\cite{defferrard2020deepsphere}. The icosahedral grids consist of equiangular triangles that form a convex polygon such that the triangles are formed by equally spaced grid points on a sphere. The number of grid points is defined by their level, with the level-0 grid being an icosahedron, and the number of grid points increasing as demonstrated in Equation \ref{eqn:icosahedron}:

\begin{equation}
\label{eqn:icosahedron}
N = 10\times2^{2g} + 2
\end{equation}

where $g$ refers to the desired grid level, and $N$ represents the number of points in the grid that form the icosahedron. This re-sampled icosahedral mesh provides data with uniform density across the sphere. In this study, we utilize a level-5 icosahedral grid, equivalent to a 1-D vector of size 10242 with a resolution of approximately 220 $km$. We employ the PyGSP library in Python to perform the grid transformation. This library is widely utilized for various graph operations in signal processing and social network analysis, such as the Erdős-Rényi network~\cite{rozemberczki2020api}. To convert to spherical grid, we first develop of a backbone structure for the icosahedral coordinate system, where the properties of the spherical coordinates (levels), are specified as inputs. At this stage, the coordinates are represented as graph networks. Next, latitude and longitude values are assigned to the graph network points $(x,~y)$ such that they can be expressed in a geographical coordinate system. Since the points in the icosahedral backbone do not exactly align with the positions in the 2D gridded Earth System Model data, we use bilinear interpolation to interpolate the ESM data with the icosahedral backbone, obtaining the final spherically-sampled data. Figure~\ref{fig:spherical_sampling} shows a schematic of grid transformation in the icosahedral spherical sampling process.

\subsubsection{Data-preprocessing}
Prior to use by AiBEDO, the CESM2 LE input and output data are preprocessed by subtracting the ensemble mean for each month and grid point in the data set. This removes the seasonal cycle of the variables, such that we consider the month-to-month deviations from the climatology, and removes the secular trend effects due to external forcings, such that we only consider internal climate variability. As we are using a LE ensemble mean, we are able to filter out the forced signal across all time scales, including short-term fluctuations such as anthropogenic aerosol forcing and volcanic eruptions \cite{rodgers_ubiquity_2021}. 

\subsection{Data-driven Models of AiBEDO}

We develop a function $A_{\tau}$ to map global perturbation of cloud at time t (input: $\delta f_t \in \mathbb{R}^{d \times c_{in}}$ ) to corresponding climate response with time-lag $l$ (output $\delta \left<y_{t+l}\right>\in \mathbb{R}^{d \times c_{out}}$).  To tackle this, we formulate the problem as a pixel-wise regression problem, learning a mapping from input to output, $A^{l}: \mathbb{R}^{d \times c_{in}} \rightarrow \mathbb{R}^{d \times c_{out}}$. We develop separate models for different time-lags from 0 (simultaneous) to 6 months at monthly intervals, and estimate climate response $\delta \left<y\right>$ using a truncated version of  equation \ref{eq:fdt_aibedo} with $\tau_{max} = 6$ months. 

\begin{figure}[H]
\begin{center}
\includegraphics[width=7.5cm]{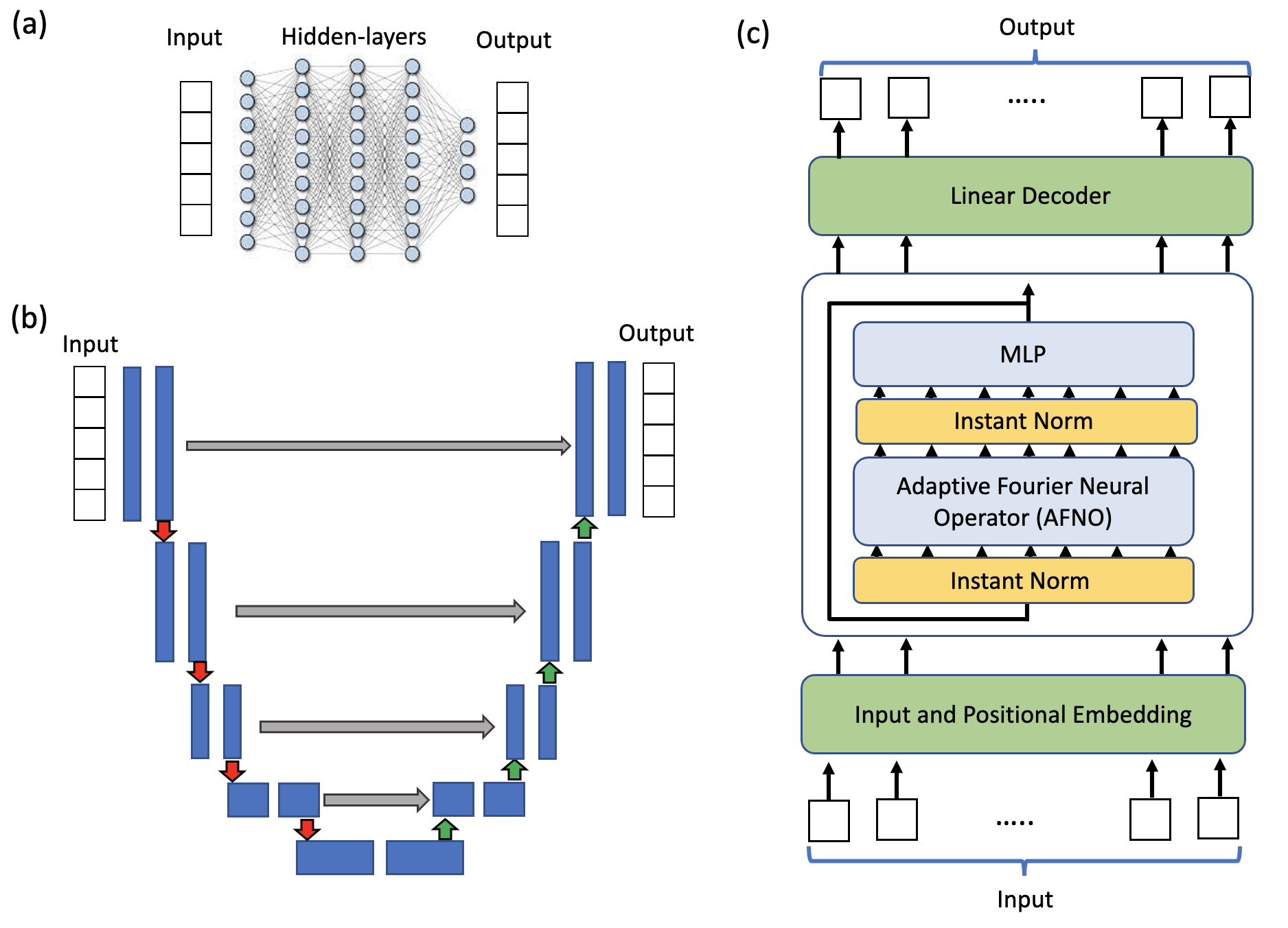}
\end{center}
\caption{Schematic view of three machine learning methods utilized for AiBEDO model, $A^l$: (a) Multilayer Perceptrons (MLPs), (b) U-Net, (c) Adaptive Fourier Neural Operator (AFNO).}
\label{fig:ml}
\end{figure}

We utilize three machine learning methods to model $A_{\tau}$ which are proven to be effective for spatio-temporal modeling of ESM data: (1) Spherical Multilayer Perceptron (S-MLP)~\cite{park2019machine, wang2014novel}, (2) Spherical U-Net (S-Unet)~\cite{mabaso2006spatio, ge2022improved, dunham2022high, trebing2021smaat} and (3) Spherical Adaptive Fourier Neural Operator (S-AFNO) ~\cite{kurth2022fourcastnet, li2020fourier}. Then, we use S-MLP for our MCB application, since S-MLP performs best out of all three methods.
We describe the three ML methods we employed further below. A schematic of the three methods is shown in Figure~\ref{fig:ml}. All machine learning models in our work are trained on a single NVIDIA Tesla V100-SXM2 GPU using 16GB VRAM. 

\subsubsection{Spherical Multilayer Perceptron (S-MLP)}:
MLP is a representative structure of Deep Neural Networks (DNNs) consisting of input and output layer inter-connected with multiple hidden layers. Each node in a layer is fully connected with all nodes in previous layer. The connection between two nodes represents a weighted value that passes through the connection signal. Each node contains a non-linear activation function to represent non-linearity of correlation between two connected nodes. The operation between consecutive layers is defined as multiplication between nodes in previous layer and corresponding weight parameters, and applying activation function. 

We combine our model architecture with spherical sampling described earlier to create Spherical Multilayer Perceptron (S-MLP). The S-MLP architecture has four hidden layers, each containing 1024 nodes. Layer normalization~\cite{ba2016layer} and Gaussian Error Linear Units (Gelu) activation~\cite{hendrycks2016gaussian} were employed in each layer. The decoupled weight decay regularization optimization method, AdamW~\cite{loshchilov2017decoupled} was utilized to train our model in an iterative manner. The learning rate was initially set to $2 \times 10^{-4}$ and exponentially decayed at a rate of $1 \times 10^{-6}$ per epoch.  We train model for 15 epochs with a batch size of 10. Our S-MLP models have $\sim$ 108M trainable parameters, and it takes around 1 minute per single epoch for training on the historical CESM2-LE dataset.

\subsubsection{Spherical U-Net (S-Unet)}:
U-Net is a symmetric U-shaped convolutional neural network for image-to-image prediction, and consists of a encoder-decoder scheme structure. The encoder extracts visual features from the input by reducing dimensions in every layer, and the decoder increases the dimensions and predict output with same size as input. Encoder and decoder are connected with long skip-connections allowing high-resolution features from the encoder are combined with the input of the decoder for better localizing visual feature in prediction. We chose to explore U-net architecture as they are generally known to capture fine spatial features in images, and are proven to do well in biomedical image segmentation~\cite{Yin2022UnetBiomed}, and satellite image analysis~\cite{McGlinchy2019UnetSatellite}.

We built our S-Unet Autoencoder architecture based on the DeepSphere model proposed by Defferrard et al. (2020)\cite{defferrard2020deepsphere}. The decoder and encoder of the S-Unet were comprised of six Chebyshev Graph Convolutional Layers\cite{boyaci2022cyberattack}, followed by spherical Chebyshev pooling, which performs spherical convolution on 1-D data considering the icosahedral geometry of the graph structure. The kernel size was set to [64,128,256,512,512,512] for the encoder and decoder, respectively. The output of the model was processed through a softmax activation. The S-Unet was trained using the AdamW optimization method~\cite{loshchilov2017decoupled}, with a learning rate of $5 \times 10^{-4}$ that was exponentially decayed at a rate of $1 \times 10^{-6}$ per epoch. The model was trained for 30 epochs, and had approximately 5.8 million trainable parameters. The training of the S-Unet model took approximately 1.5 minutes per epoch in our computing environment.  


\subsubsection{Spherical Adaptive Fourier Neural Operator (S-AFNO):}
Motivated by the recent successes of transformer-based architectures in climate domain, we adopt the FourcastNet\cite{kurth2022fourcastnet} employing Adaptive Fourier Neural Operator (AFNO)~\cite{guibas2021efficient} to tackle our problem.
FourcastNet is a weather forecasting model using the multi-layer transformer architecture employing AFNO inside. The input is first divided into multiple patches which are embedded in a higher dimensional space
with larger number of latent channels and corresponding positional embeddings. Then, positional embeddings are formulated as the sequence of tokens. Tokens are spatially mixed using AFNO followed by subsequent mixing of latent channels accordingly. Mixed embeddings are passed by MLP to learn higher level feature. We repeat this process for each transformer layer. After this process, a linear decoder reconstruct the patches from final embedding.  

Instead of the token projection technique employed in ForecastNet, which involves the composition of tokens from a two-dimensional grid patch of climatic data, we project all elements of a spherically resampled one-dimensional input as tokens. These tokens, together with a positional encoding, are then inputted into a sequence of AFNO layers. Each layer, upon receiving an input tensor of tokens, performs spatial mixing followed by channel mixing. We name our model Spherical Adaptive Fourier Neural Operator (S-AFNO). For S-AFNO model, we used 4-layered transformer~\cite{vaswani2017attention}, and set the size of embedding of token as 384. We use GeLu activation\cite{hendrycks2016gaussian} for the output of MLP layers while applying layer normalization techniques~\cite{ba2016layer} to stabilize training. The S-AFNO model was trained using the AdamW optimization ~\cite{loshchilov2017decoupled}, with a learning rate of $5 \times 10^{-4}$ that was exponentially decayed at a rate of $1 \times 10^{-6}$ per epoch. The model was trained for 50 epochs, and had approximately 9 million trainable parameters. The training of the S-AFNO model took approximately 12 minutes per epoch using CESM2-LE training data in our computing environment. 

\begin{figure}[H]
\begin{center}
\includegraphics[width=10cm]{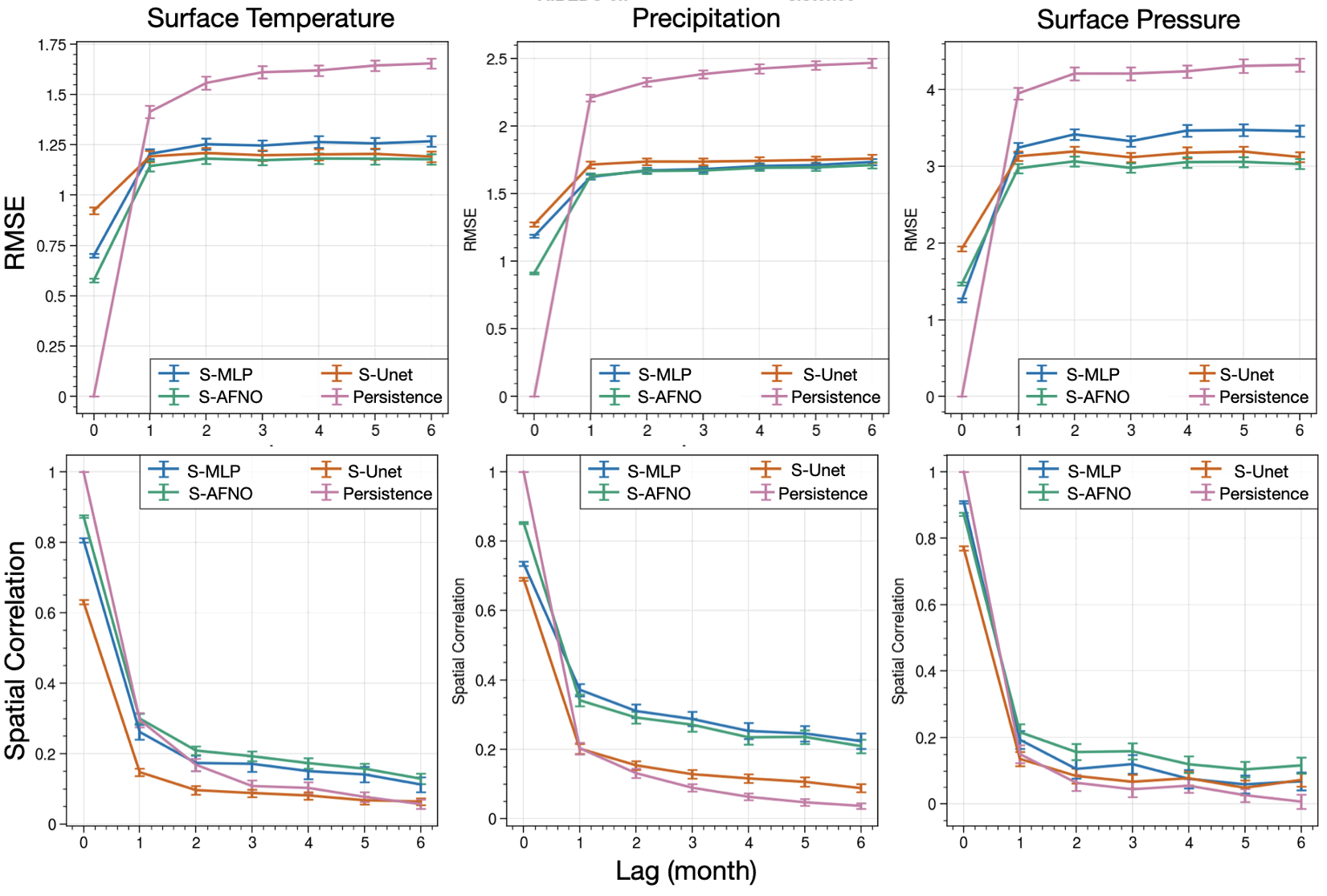}
\end{center}
\caption{The performance of AiBEDO in emulating CESM2-LE data using different ML models including S-MLP, S-Unet, S-AFNO. The persistence curve illustrates the temporal deviation of the reference output, quantified as the discrepancy between the ground-truth output at each lag compared with no lag (when lag = 0).}
\label{fig:aibedo_emulation}
\end{figure}

\begin{figure}[H]
    \centering
    \includegraphics[width=\columnwidth]{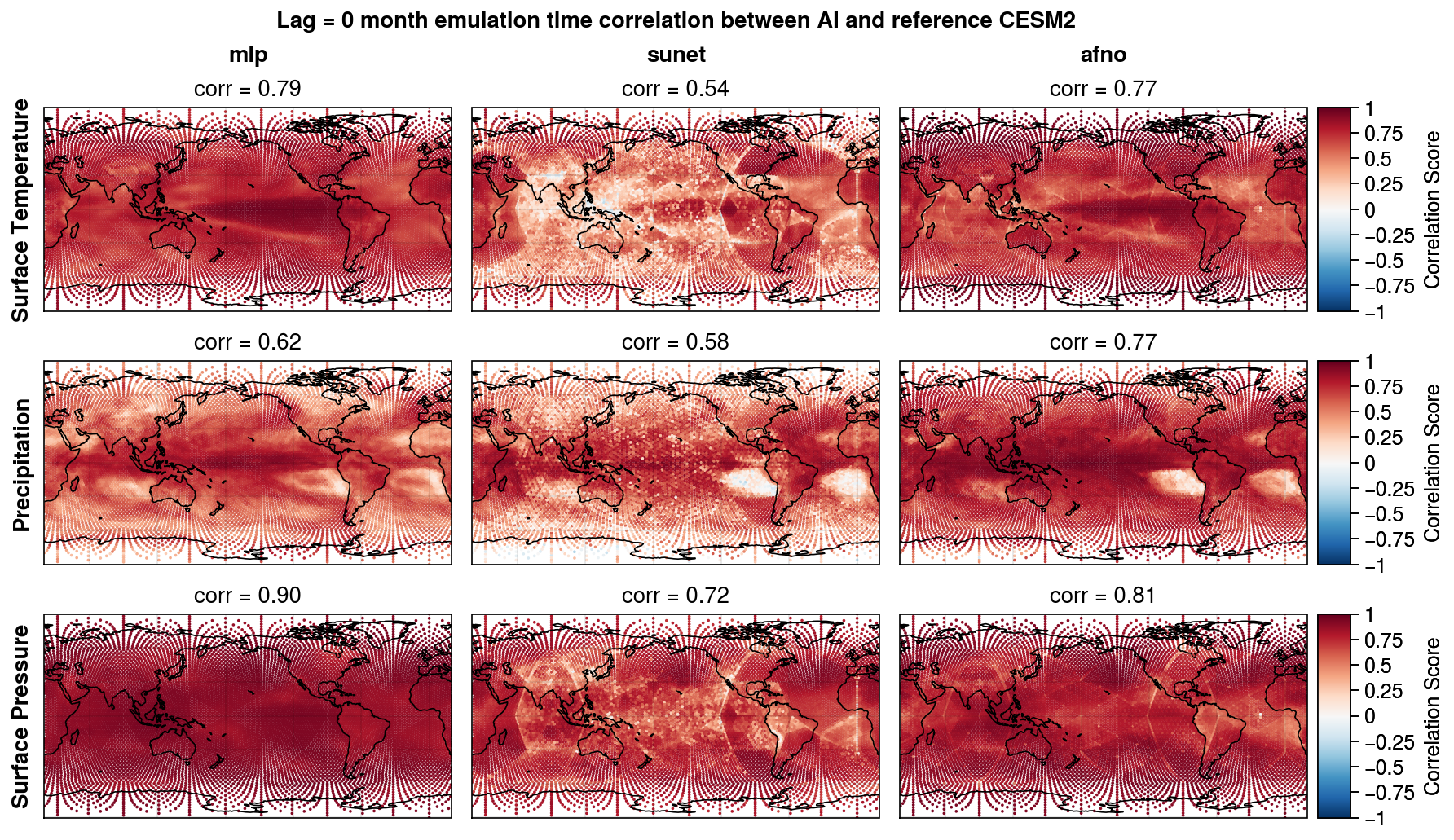}
    \caption{Time correlation scores of simultaneous (0-lag) AiBEDO S-MLP (left column), S-Unet (middle column), and S-AFNO (right column) emulation versus baseline CESM2 data for surface temperature (top row), precipitation (middle row), and surface pressure (bottom row).}
    \label{fig:lag0_corrmaps}
\end{figure}

\begin{figure}[H]
    \centering
    \includegraphics[width=\columnwidth]{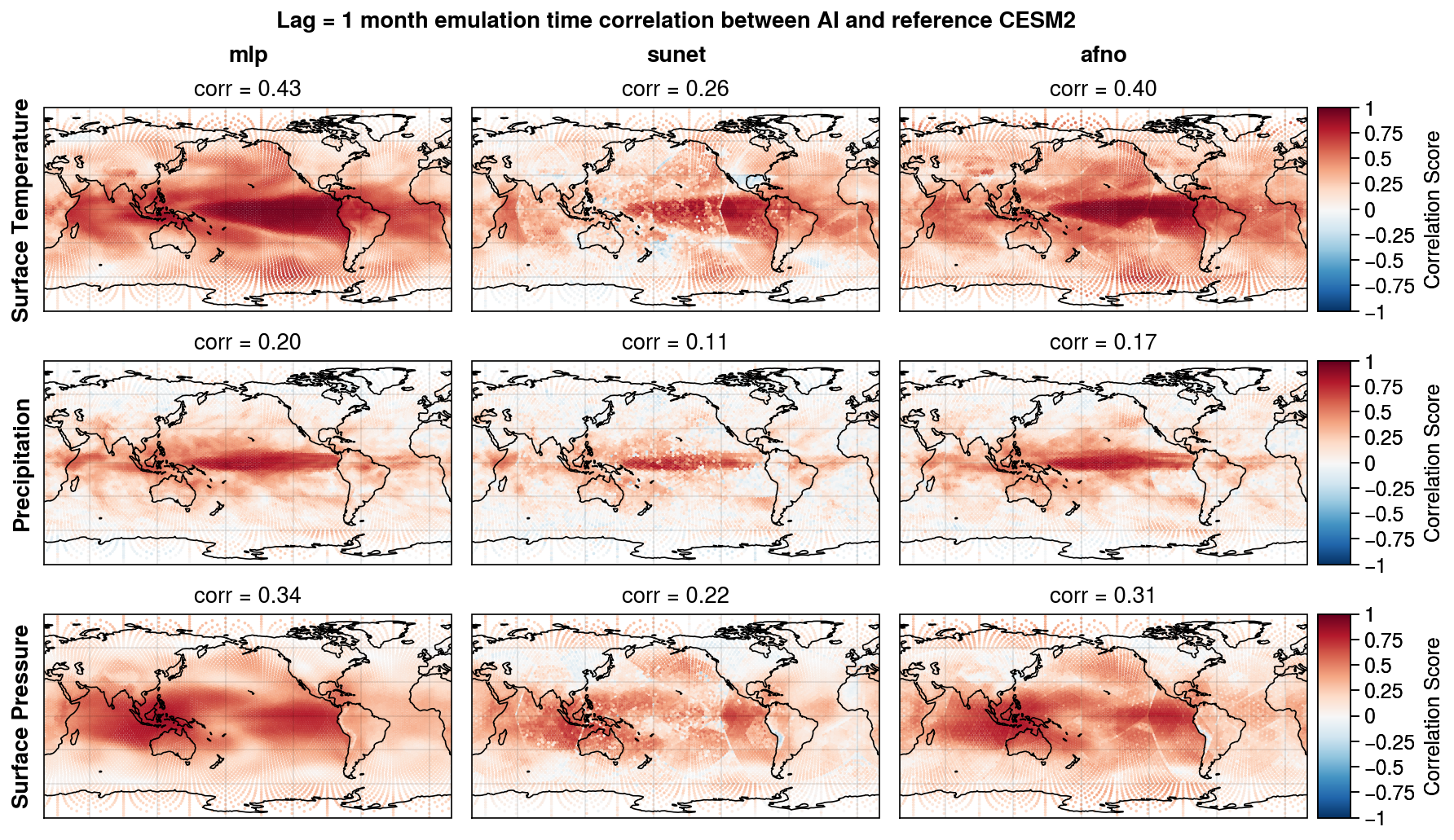}
    \caption{Time correlation scores of one-month lag AiBEDO S-MLP (left column), S-Unet (middle column), and S-AFNO (right column) emulation versus baseline CESM2 data for surface temperature (top row), precipitation (middle row), and surface pressure (bottom row).}
    \label{fig:lag1_corrmaps}
\end{figure}

\subsection{AiBEDO Marine Cloud Brightening Experiments}

To evaluate how well AiBEDO estimates climate responses, we compare the impact of MCB-like forcing in the lag-integrated AiBEDO response to fully coupled CESM2 simulations \cite{hirasawa_impact_nodate}. These simulations are summarized in Table \ref{tab:cesm2} in the Appendix along with a description of how the MCB forcing perturbations are computed from CESM2 for AiBEDO. Our experiments focus on perturbing three main regions (shown in Figure \ref{fig:regpert_tas}) for MCB experiments: NEP (North East Pacific - 0 to 30N and 150W to 110W), SEP (South East Pacific - 30S to 0 and 110W to 70W) and SEA (South East Atlantic - 30S to 0 and 15W to 25E) regions. Following equation \ref{eq:fdt_aibedo}, for each lag we calculate the AiBEDO response by taking the mean of the difference between the AiBEDO output for an $N=480$-month sample of CESM2 internal variability perturbed with the MCB radiation anomalies~\cite{aswathy2015climate} minus the AiBEDO output for that same 480-month sample without perturbations. We use this protocol rather than simply running AiBEDO with the radiation anomaly fields, because the near-zero anomalies outside the perturbation regions cause artifacts in the AiBEDO output, as such a field is entirely unlike any fields the model is trained on. 


\section{Results and Discussion}

We evaluate model functionality through two phases of AiBEDO development. The first phase involves developing an AI emulator of simultaneous and time-lagged mappings. As described in the previous sections, we developed three AI models (S-MLP, S-UNet, S-AFNO) and compared their performance in how well they could emulate the mappings. The second phase encompasses an analysis of the results from MCB experiments conducted in selected regions on CESM2, along with the time-integrated outputs generated by AiBEDO through the application of an FDT-like function. We also compared the time-integrated results across the three AI models to analyze the performance of different model architectures. 

\subsection{Comparison between AiBEDO ML algorithms (Phase I)}

Each machine learning model was trained individually with different monthly time-lags ranging from 0 to 6 months. The quantitative comparison of emulation performance of AiBEDO using different ML models are reported in Figure \ref{fig:aibedo_emulation}. We evaluate Root Mean Squared Error (RMSE) and spatial correlation score of prediction with ground truth of our models based on the persistence scores of the ground truth data. \textit{Persistence} at a time-lag $t$ is essentially the temporal deviation of the ground truth outputs at $t$ months from the zero time-lag instances. The persistence curve at different time-lags is overlaid with performance curve from our model in  Figure~\ref{fig:aibedo_emulation}, which lets us evaluate the ML models performance over lagged duration. As shown in Figure \ref{fig:aibedo_emulation}, we can observe that as the time-lag increases, the predictive accuracy of the model decreases as anticipated. Notably, the model consistently outperforms persistence across all time-lags, implying that AiBEDO has learned the temporal dynamics patterns beyond the simple memory of 0-month temperature anomalies.

Figures \ref{fig:lag0_corrmaps} and \ref{fig:lag1_corrmaps} demonstrate the spatial correlation scores of the emulation of each ML model (S-MLP, S-Unet and S-AFNO) for simultaneous and one-month time-lagged model.  The corresponding qualitative comparison of sample timesteps is shown in Figures \ref{fig:aibedo_emulation_lag0} and \ref{fig:aibedo_emulation_lag1} in the Appendix. Despite S-AFNO achieving comparable score to S-MLP in terms of RMSE and spatial correlation for simultaneous AiBEDO results, as demonstrated in Figures ~\ref{fig:lag0_corrmaps} and \ref{fig:aibedo_emulation_lag0}, S-MLP was found to outperform the other models in capturing global response patterns in all three variables. The quality deviation of output between the models becomes evident in the time-lagged response even after a single month (Figure ~\ref{fig:lag1_corrmaps}), showing higher performance of S-MLP. The qualitative results with a time-lag demonstrate that S-MLP significantly outperforms the other two models, S-AFNO and S-Unet, even though RMSE scores are comparable across the models.

The superior performance of S-MLP model can be attributed to its significantly larger number of trainable parameters and its fully connected network structure. This allows the model to consider the correlation of each element in the input, thereby enhancing its ability to regress the results. The fully connected network structure enables the output to be regressed by multiplying the different weights of all input elements, which enhances its ability to capture long-range interactions between locations, potentially driven by the ``butterfly effect''. The subpar performance of the S-Unet can be attributed to the loss of important information regarding global climate patterns, particularly related to long-range interactions, during the spherical convolution and pooling processes. Consequently, machine learning models based on the assumption of spatial locality, such as the S-Unet, underperform models that allow for global attention. While S-AFNO performed better than S-Unet, the model still underperforms compared to S-MLP. The amount of training data used in this study may not be sufficient for training a transformer-based S-AFNO model, which typically requires a large amount of data. Moreover, the time taken to complete one epoch in S-AFNO was already significantly higher than time per epoch for S-MLP. Adding more parameters and/or data can only increase the computational requirement for S-AFNO. As a result, S-MLP was utilized in the subsequent experiments described in this paper.

\subsection{Response to MCB perturbations (Phase II)}
\begin{figure}[H]
\begin{center}
\includegraphics[width=\linewidth]{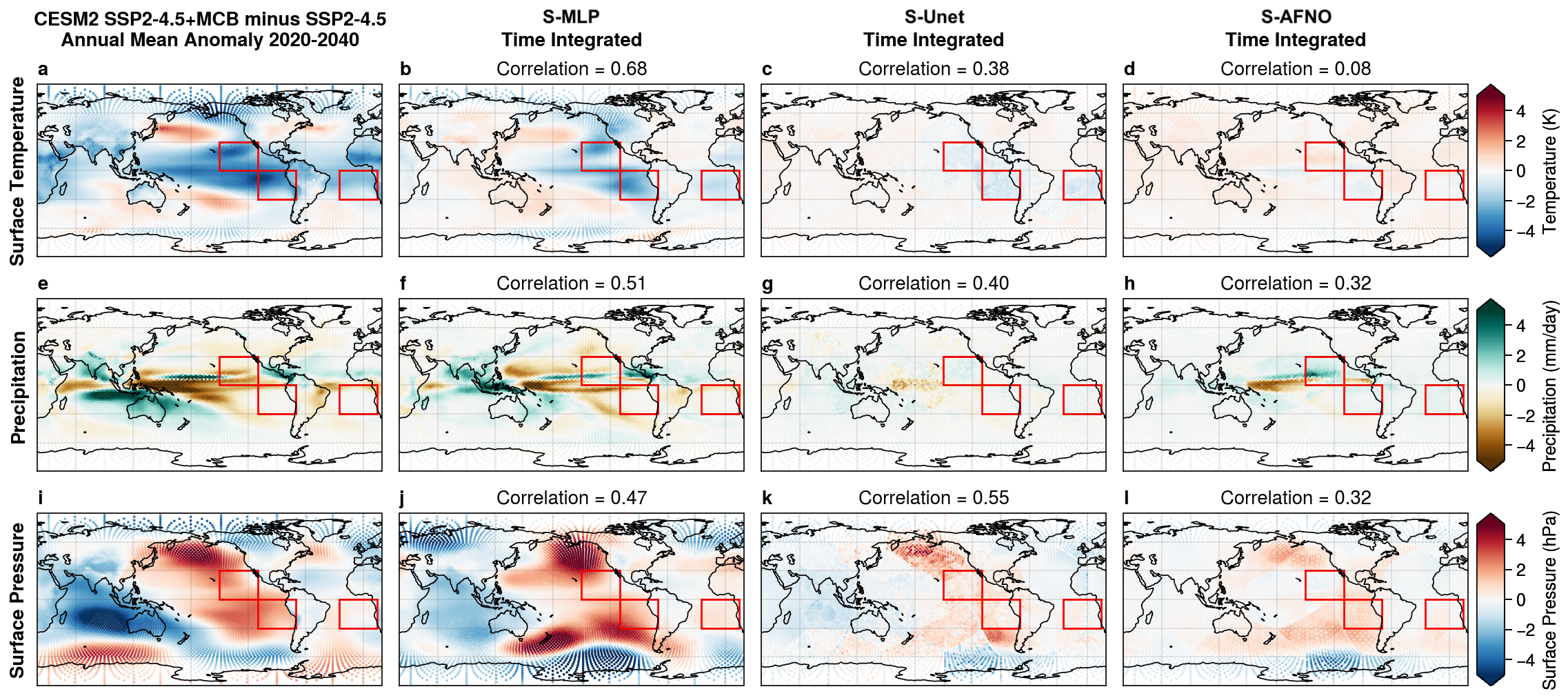}
\end{center}
\caption{Comparison of MCB Perturbation Responses in 6-month integrated S-MLP, S-Unet and S-AFNO Models to CESM2 coupled simulations.}
\label{fig:aibedo_mcb_comparison}
\end{figure}

\begin{figure}[H]
\centering
\includegraphics[width=1\columnwidth]{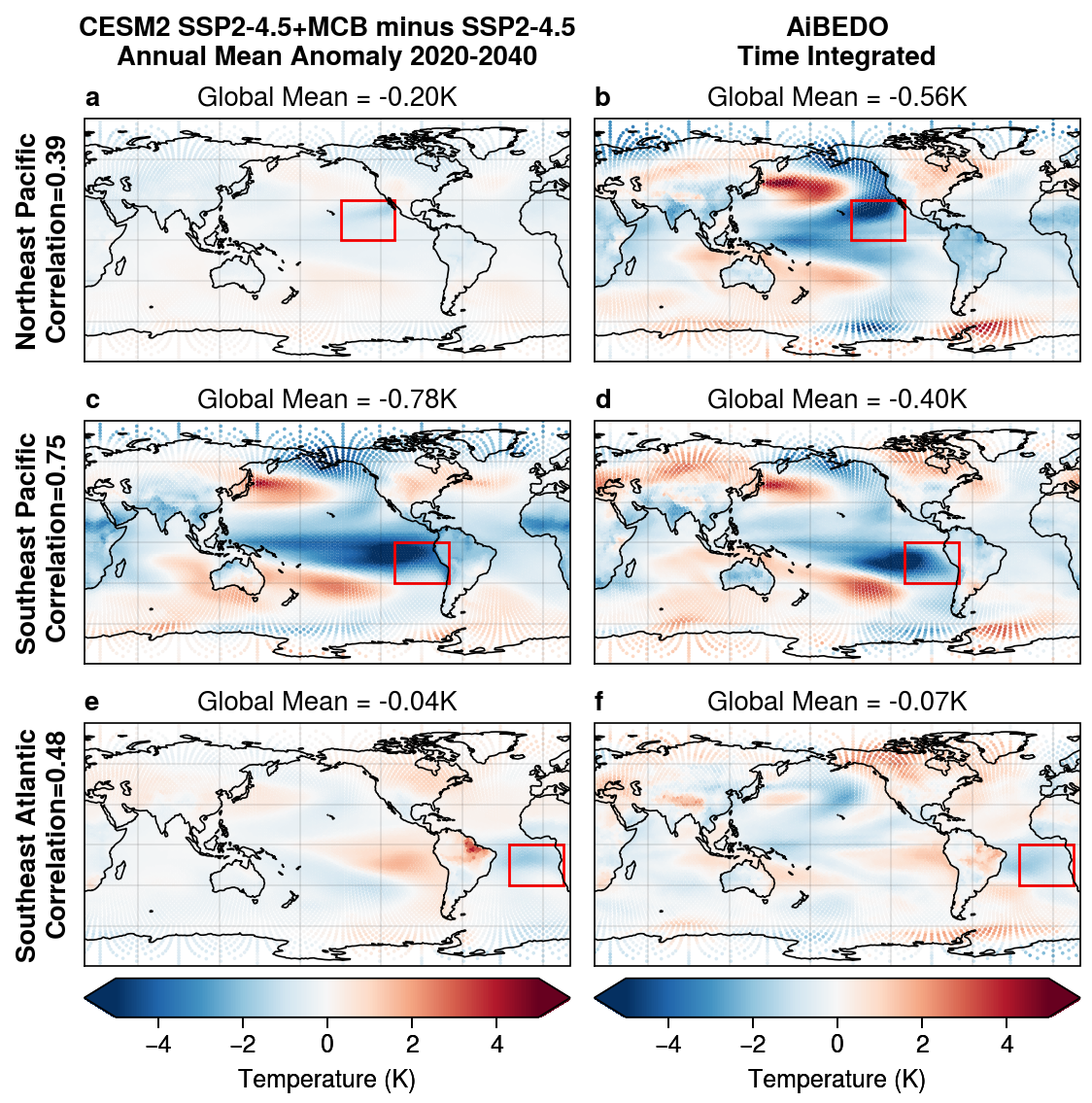} 
\caption{CESM2 (left column) and lag-integrated AiBEDO (right column) surface temperature anomalies due to MCB-like forcing in the NEP (top row), SEP (middle row), and SEA (bottom row). AiBEDO responses are computed using Simpson's rule integration of anomalies from lag 1, 2, 3, 4, 5, 6, 12, 24, 36, and 48 month models.}
\label{fig:regpert_tas}
\end{figure}

Here, we compare the impact of MCB-like forcing in three regions in the subtropics on the climate in CESM2 and lag-integrated AiBEDO. Figure \ref{fig:regpert_tas} shows the spatial maps of the surface temperature anomalies for the response to each of the three regions (NEP, SEP, and SEA). The goal of comparing the response of the CESM2 model and the AiBEDO model to MCB-like perturbations is to determine if AiBEDO can accurately estimate the forced climate response. Figure \ref{fig:aibedo_mcb_comparison} shows that the S-MLP model within AiBEDO has better performance compared to the other two models, reproducing the climate response pattern with a correlation score of 0.68 for temperature (\texttt{tas}), 0.51 for precipitation (\texttt{pr}), and 0.47 for pressure (\texttt{ps}). The correlation score improves with increased time-integration. However, there are differences in the magnitude of the responses, which is likely due to missing lags in the integration.

The S-MLP model successfully estimates remote teleconnected responses to MCB forcing, such as a La Niña-like temperature signal in the Pacific. It also captures drying in northeast Brazil, central Africa, and southern North America and Europe, and wetting in the Sahel, south and southeast Asia, Australia, and central America. The S-Unet and S-AFNO models capture some aspects of the responses, such as tropical Pacific drying and high pressure anomalies in the Pacific mid-latitudes. However, they miss many key aspects of the response patterns. These responses can be used to assess the risk of regional tipping points, such as Amazon dieback and Sahel region in North Africa greening \cite{mckay_exceeding_2022} or the reduction in the risk of coral dieoff due to cooling in the tropical oceans. However, due to the lower performance of AiBEDO at high latitudes, it may be challenging to evaluate key cryospheric tipping points such as permafrost loss in Eurasia and North America.

The impact of MCB forcing on individual perturbation regions is also evaluated and compared to CESM2 simulations with similar regional forcing. Figure \ref{fig:regpert_tas} shows the comparison of perturbation effects on temperature between CESM2 and S-MLP model of AiBEDO. In general, AiBEDO's performance is weaker when considering these regional perturbations compared to when all three regions are perturbed together. AiBEDO performs best for SEP, followed by SEA and NEP. The NEP response is overestimated substantially, which may indicate the model learns too heavily from the El Niño-Southern Oscillation climate pattern (variations in ocean temperatures in the Pacific Ocean) at the expense of over modes of variability. However, AiBEDO correctly identifies the climate responses to the different forcing regions in several key regions. For example, it correctly identifies that SEP forcing causes La Niña-like cooling (below-average sea surface temperatures in the central and eastern tropical Pacific Ocean) and that SEA forcing causes tropical Pacific warming and Amazon drying. Moreover, AiBEDO performs better in the tropics and over oceans compared to higher latitudes and over land. This aligns with the regions where AiBEDO has the highest emulation skill, indicating that the model's ability to correctly estimate climate responses to MCB forcing is closely linked to its ability to emulate internal variability.


\subsection{Visualization Platform}

\begin{figure}[H]
\centering
\includegraphics[width=15cm]{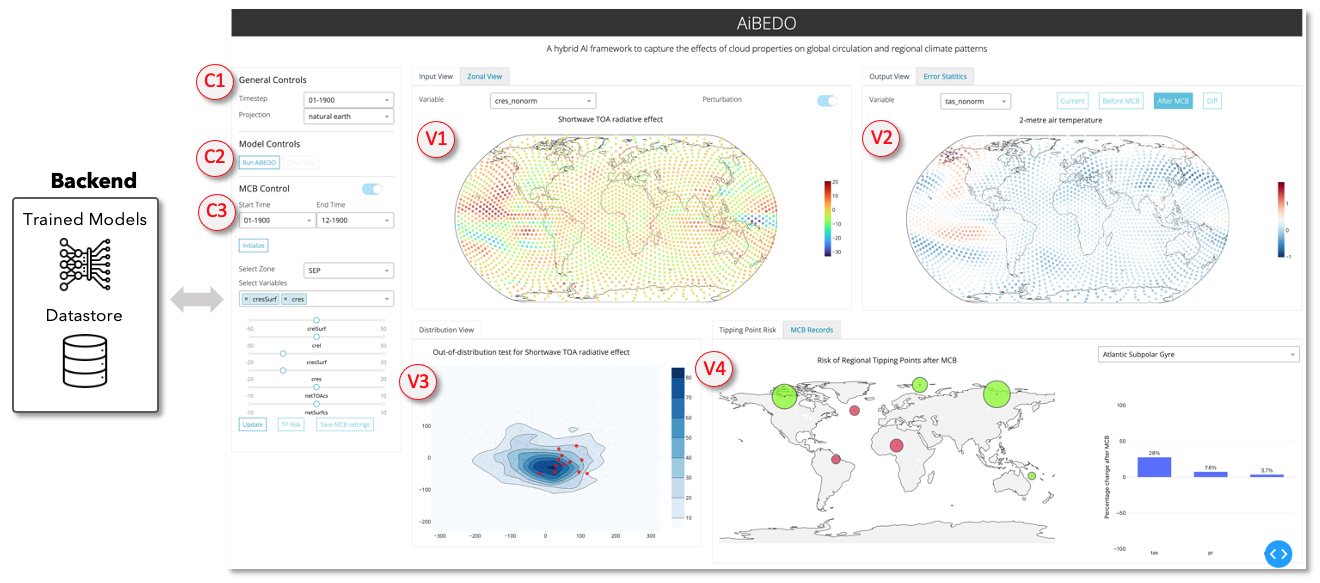} 
\caption{High-level overview of our interactive visual analysis system to work with the trained hybrid models and drive post-hoc analysis for different MCB scenarios. [source code:\url{https://github.com/subhashis/aibedoviz}, demo video:\url{https://youtu.be/3dmqYqkSLOo}]  }
\label{fig:viz_overview}
\end{figure}

We developed a front-end interactive visualization platform to facilitate downstream tasks and exploratory analyses using the suite of trained hybrid AI models. It allows climate scientists to interact directly with the trained models (described above) and recreate different MCB experiment scenarios. The multi-panel design lets the experts load different Earth System Model data, interactively run the trained hybrid model in the backend and visualize the model predictions and inputs using popular geospatial projection schemes. Figure~\ref{fig:viz_overview} shows an overview of our visualization platform. The leftmost panel in Figure~\ref{fig:viz_overview} is the general control panel which permits a subject matter expert to interact with the models and data source, as well as tune several experiment parameter for analysis. Figure~\ref{fig:viz_overview}($C_1$) highlights the controls to specify the specific timestamp from the data and the general visualization projection scheme. Figure~\ref{fig:viz_overview}($C_2$) corresponds to the hybrid model controls, which lets us run the trained AI models with specified data directly from this interface, as well as clear any previous model prediction results from memory. Figure~\ref{fig:viz_overview}($C_3$) provides the main MCB experiment scenario controls. It helps to set the important parameters like which geospatial regions to apply MCB over, which input variable set to perturb and by what extend and running the perturbed data with the AI models. Figure~\ref{fig:viz_overview}($V_1$) and ($V_2$) highlight the input and output visualization panels respectively. Using the set geospatial projection scheme (as set in $C_1$) the different input and output variables are visualized here. In addition, the output panel has the options to show the original AI model prediction results, results after MCB, and the net difference in results as well. This helps to qualitatively analyze the output predictions and their likely patterns.

Since the perturbed input fields prepared as part of the MCB experiments may lie outside the scope of the input distribution with which the hybrid models were trained over, it is important to check the out-of-distribution cases. To address this, in Figure~\ref{fig:viz_overview}($V_3$) we visualize the low-dimensional projection of the MCB perturbed fields over the input distribution to show how far off the perturbed fields are from the original data. Another important aspect of our analysis is to understand the impact of MCB on key regional tipping points. Figure~\ref{fig:viz_overview}($V_4$) panel shows the seven regional climate tipping points that we are tracking with each MCB experiment scenarios. A red color on this sites indicates that the current MCB setting might trigger factors that are directly associated with the risk of tipping points in these sites.

\section{Summary and Future Work}

In this study, we introduce a novel methodology for utilizing the Fluctuation-Dissipation Theorem, a principle derived from statistical mechanics, in the context of  knowledge discovery using AI model to predict the behavior of a system under external forcing. This approach is particularly useful in scenario analysis in scientific domains where traditional models are computationally infeasible. We demonstrate the efficacy of this method through its application in Marine Cloud Brightening (MCB) climate intervention analysis, where the number of scenarios required is intractable using traditional computational models. This constitutes a challenging problem in climate science, where Earth System Models are employed to simulate changes in climate and require weeks to complete a single run, and assessing additional scenarios necessitates additional runs. Our method of training lagged emulation models of internal variability (noise) and producing time-integrated outputs using an FDT-like framework presents a viable alternative for fast prototyping of such scenarios in several scientific domains. In addition to the central contribution of our work, we also present various components of AiBEDO that is specific to climate intervention: preprocessing techniques for large climate datasets, a comparison of machine learning methods for time-integrated approach, and a user-friendly visualization interface that provides explainable insights that helps with model investigation and tipping point analysis, and allows for quick what-if scenario analysis.

Our next steps include expanding the MCB climate intervention framework into developing optimal pathways for selected tipping point scenarios. Figure \ref{fig:optimize_mcb} in Appendix shows some of the known tipping points in the climate system and the associated mitigating strategies that may be used to avoid them. An inverse search of AiBEDO output may be useful for creating MCB perturbations to achieve such mitigation measures. For example, to avoid the catastrophic retreat of the West Antarctic ice sheet, potential MCB perturbation sites could push the jet circulation equatorward. To achieve this, the user might query the model to create a forcing scheme to avoid selected tipping points, and the model would choose optimal spatiotemporal perturbation MCB sites that could lead to this outcome. We could also add additional constraints not to tip other climate tipping points in this process. This is a powerful tool that could help scientists and policymakers understand the climate system's teleconnections that could trigger unintended consequences and timely strategies to avoid catastrophic outcomes due to changing climate.

\section{Funding and Acknowledgements}

The development of AiBEDO is funded under the DARPA AI-assisted Climate Tipping-point Modeling (ACTM) program under award DARPA-PA-21-04-02. The authors are grateful for the donation of computing credits and storage from AWS, which enabled the implementation of MCB scenarios utilizing CESM2 Earth Systems Model and the training of the AiBEDO model on the cloud.

\newpage
\bibliography{sample-base}

\newpage
\section{Appendix}

\subsection{Emulation of Climate Noise}
\begin{figure}[H]
\begin{center}
\includegraphics[width=\linewidth]{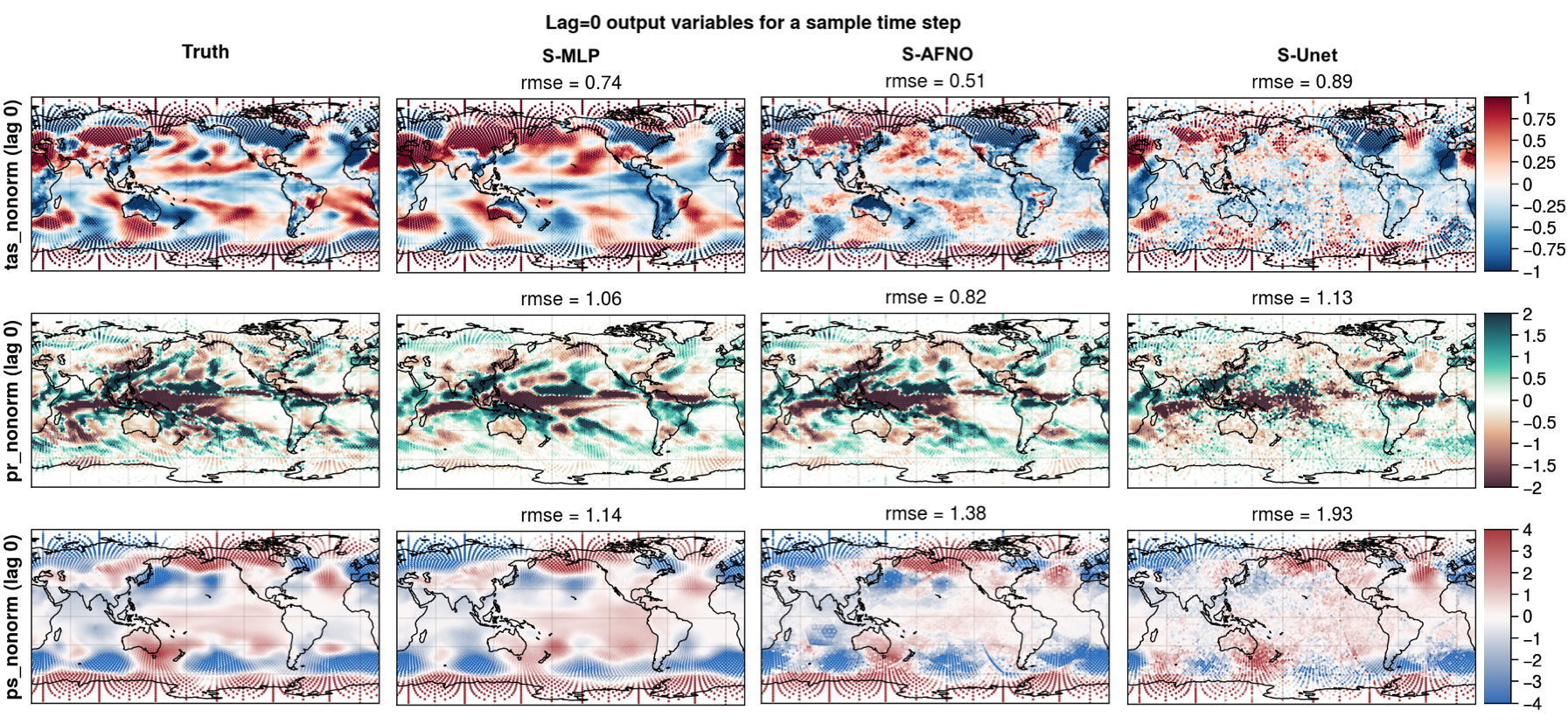}
\end{center}
\caption{AiBEDO emulation results for a sample input time step with no time-lag (simultaneous run) for surface temperature (tas), precipitation (pr), and surface pressure (ps) compared with ground truth (first columns) and different ML models. }
\label{fig:aibedo_emulation_lag0}
\end{figure}

\begin{figure}[H]
\begin{center}
\includegraphics[width=\linewidth]{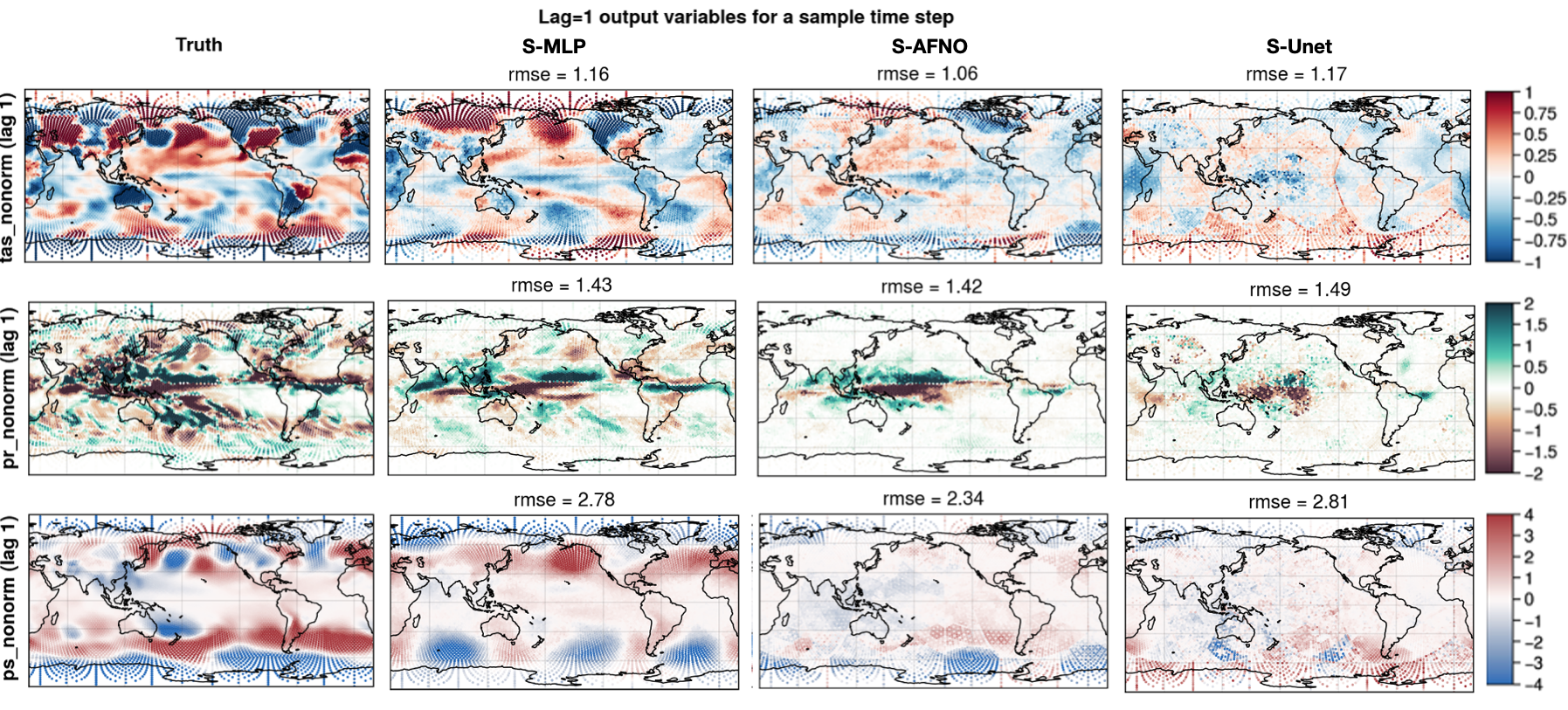}
\end{center}
\caption{AiBEDO emulation results for a sample input time step with time-lag after one month (lag =1) for surface temperature (tas), precipitation (pr), and surface pressure (ps) compared with ground truth (first columns) and different ML models. }
\label{fig:aibedo_emulation_lag1}
\end{figure}

AiBEDO emulation performance is evaluated by taking the difference AiBEDO outputs at a given lag to the corresponding preprocessed CESM2 output field at the same lag. Example emulation outputs for a single input time step are shown for the lag-0 (Figure \ref{fig:aibedo_emulation_lag0}) and lag-1 (Figure \ref{fig:aibedo_emulation_lag1}). 

\subsection{MCB Experiment Design}
\label{appendix:CESM2_MCB}
In addition to the CESM2 LE data used to train the model, we use a set of new CESM2 LE simulations that simulate the impact of MCB on climate \cite{hirasawa_impact_nodate} (Table \ref{tab:cesm2}). MCB forcing is imposed by prescribing in-cloud liquid cloud droplet number concentrations (similar to \citep{rasch_geoengineering_2009}, \citep{jones_climate_2009} and \citep{stjern2018response}) to 600cm$^{-3}$ in three selected regions in the northeast Pacific, southeast Pacific, and southeast Atlantic. We apply the MCB perturbations in the three regions separately and all together from 2015 to 2065 against a background Shared Socioeconomic Pathway 2 - 4.5Wm$^{-2}$ (SSP2-4.5) scenario. The CESM2 MCB climate response is thus computed as the difference between SSP2-4.5 plus MCB simulations minus baseline SSP2-4.5 (Shared Socioeconomic Pathway 2 - 4.5Wm$^{-2}$ forcing) simulations.

\begin{figure}[H]
\centering
\includegraphics[width=12cm]{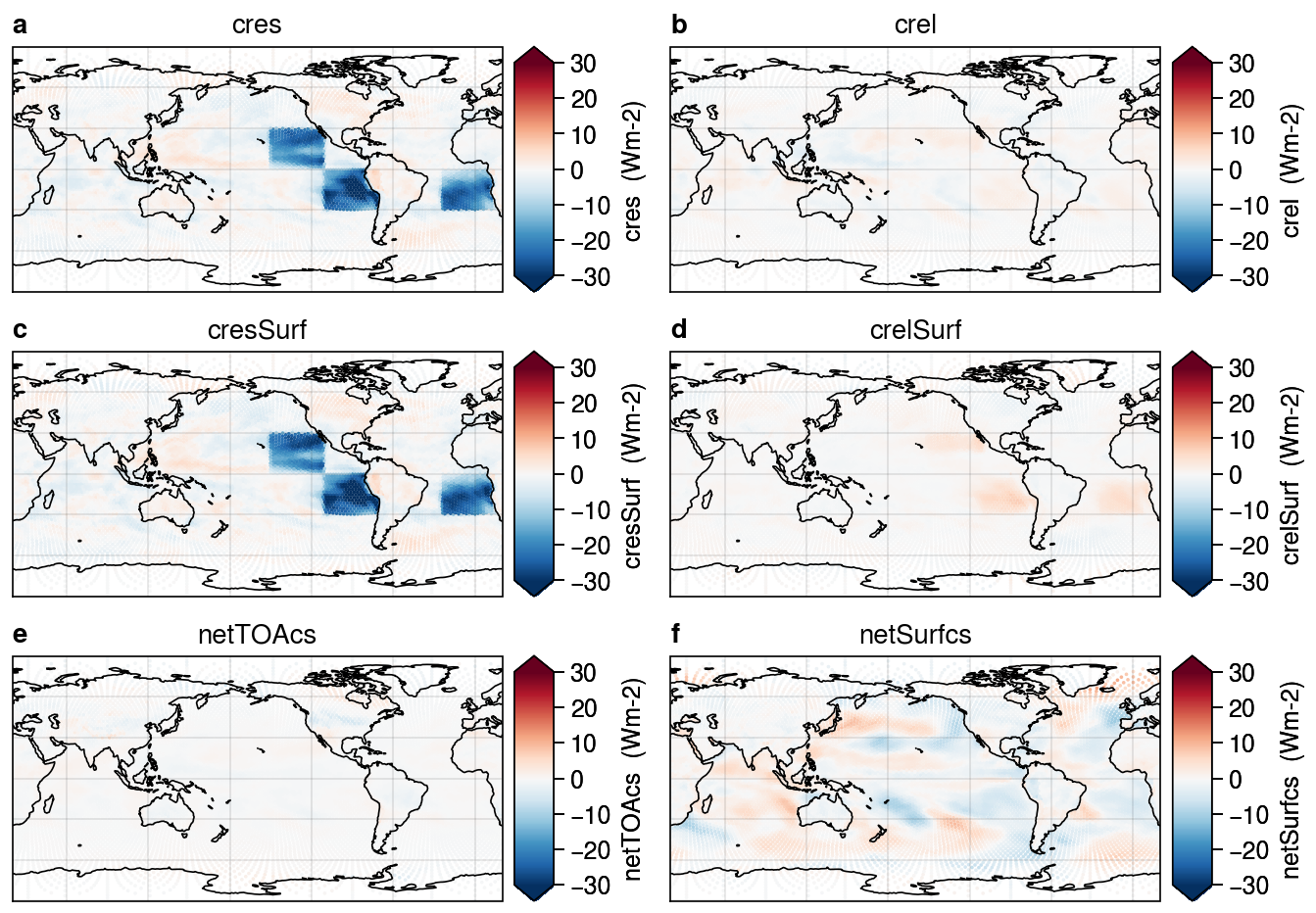} 
\caption{Annual mean radiation anomalies calculated from Fixed SST simulations and applied as MCB perturbations to AiBEDO.}
\label{fig:mcb_forcing}
\end{figure}

To compute the radiation perturbations for AiBEDO MCB responses, we use CESM2 ``fixed-sea surface temperature" (fixed SST) simulations. In these simulations, a MCB forcing identical to that used in the coupled simulations is imposed in the model with SSTs held to climatological values. This allows the computation of the radiation anomalies in the absence of any radiative feedbacks due to surface temperature change, which is referred to as the effective radiative forcing (ERF). Thus, we can compare CESM2 and AiBEDO responses for the same perturbation fields. AiBEDO perturbation fields are thus calculated as the annual mean anomalies in \texttt{cres}, \texttt{crel}, \texttt{cresSurf}, \texttt{crelSurf}, \texttt{netTOAcs}, and \texttt{netSurfcs} anomaly fields (Figure \ref{fig:mcb_forcing}) between the Y2000 MCB Perturbed minus Y2000 Control simulations. 

\begin{table*}[t]
\centering
\begin{tabular}{|l|c|c|c|c|}
    \hline
    \textbf{Experiment} &\textbf{Role} & \textbf{Forcing} & \textbf{Time span} & \textbf{N}\\
    \hline 
    Historical LE & training, testing, validation & historical & 1850 - 2015 & 50 \\
    \hline \hline
    Y2000 Control & perturbation & Year 2000 Fixed SST & 1 - 20 & N/A \\
    \hline
    Y2000 MCB Perturbed & perturbation &\begin{tabular}{@{}c@{}} Year 2000 Fixed SST + \\ MCB in NEP, SEP, and SEA\end{tabular} & 1 - 10 & N/A \\
    \hline \hline
    SSP2-4.5 LE & response validation & SSP2-4.5 & 2015 - 2100 & 17 \\
    \hline
    SSP2-4.5 + ALL MCB & response validation & \begin{tabular}{@{}c@{}} SSP2-4.5 + 600cm$^{-3}$ CDNC \\ in NEP, SEP, and SEA\end{tabular} & 2015 - 2065 & 3 \\
    \hline
    SSP2-4.5 + NEP & response validation & SSP2-4.5 + 600cm$^{-3}$ CDNC in NEP & 2015 - 2065 & 3 \\
    \hline
    SSP2-4.5 + SEP & response validation & SSP2-4.5 + 600cm$^{-3}$ CDNC in SEP & 2015 - 2065 & 3 \\
    \hline
    SSP2-4.5 + SEA & response validation & SSP2-4.5 + 600cm$^{-3}$ CDNC in SEA & 2015 - 2065 & 3 \\
    \hline
\end{tabular}
\caption{CESM2 simulations used to train and verify AiBEDO. NEP, SEP, SEA denote regions where MCB forcing is imposed, where NEP - Northeast Pacific (0 to 30N; 150W to 110W), SEP - Southeast Pacific (30S to 0; 110W to 70W), SEA - Southeast Atlantic (0 to 30N; 25W to 15E). Note the fixed SST simulations use constant climatological conditions, so we do not note specific years for these simulations.}
\label{tab:cesm2}
\end{table*}


\begin{figure}[H]
\centering
\includegraphics[width=\linewidth]{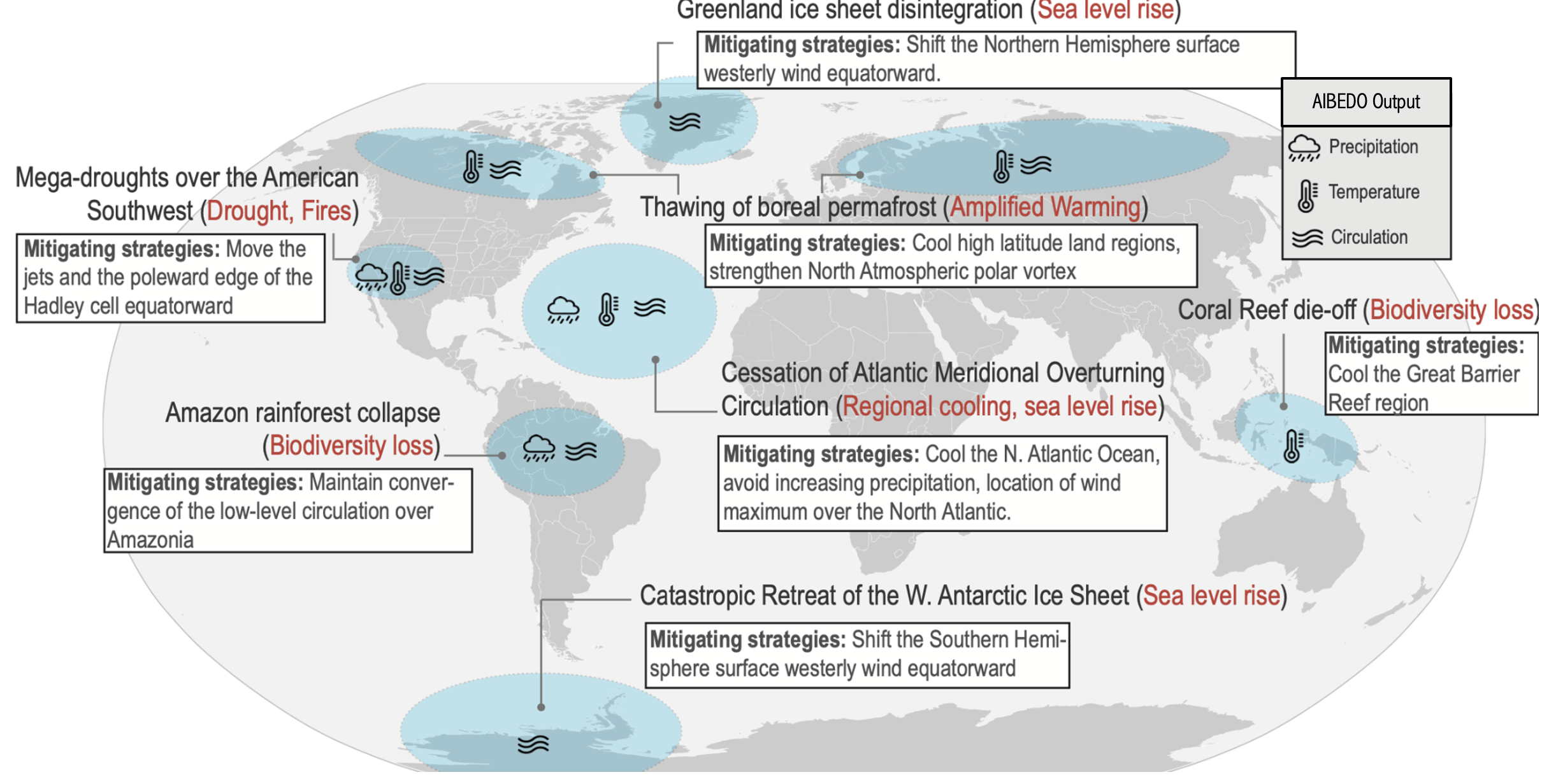} 
\caption{Climate tipping points and mitigation strategies using AiBEDO framework}
\label{fig:optimize_mcb}
\end{figure}

\end{document}